\pdfoutput=1

\documentclass[11pt]{article}

\usepackage[final]{acl}

\usepackage{times}
\usepackage{latexsym}
\usepackage{stfloats}

\usepackage[T1]{fontenc}

\usepackage[utf8]{inputenc}

\usepackage{microtype}

\usepackage{inconsolata}

\usepackage{graphicx}
\usepackage{booktabs}
\usepackage{multirow}
\usepackage{hyperref}
\usepackage{longtable}
\usepackage[noend]{algpseudocode}
\usepackage{algorithm}

\title{``Let's Argue Both Sides'': Argument Generation Can Force Small Models to Utilize Previously Inaccessible Reasoning Capabilities}

\author{Kaveh Eskandari Miandoab\textsuperscript{+} \and Vasanth Sarathy\textsuperscript{*} \\
Tufts University \\
\texttt{kaveh.eskandari\_miandoab@tufts.edu\textsuperscript{+}}  \\ \texttt{vasanth.sarathy@tufts.edu\textsuperscript{*}}
}

\usepackage{enumitem}
\usepackage{cancel}
\usepackage[draft]{commenting}
\declareauthor{vs}{Vasanth}{blue}
\declareauthor{kem}{Kaveh}{orange}

\begin{document}
\maketitle
\begin{abstract}
    Large Language Models (LLMs), despite achieving state-of-the-art results in a number of evaluation tasks, struggle to maintain their performance when logical reasoning is strictly required to correctly infer a prediction. In this work, we propose \emph{Argument Generation} as a method of forcing models to utilize their reasoning capabilities when other approaches such as chain-of-thought reasoning prove insufficient. Our method involves the generation of arguments for each possible inference result, and asking the end model to rank the generated arguments. We show that \emph{Argument Generation} can serve as an appropriate substitute for zero-shot prompting techniques without the requirement to add layers of complexity. Furthermore, we argue that knowledge-probing techniques such as chain-of-thought reasoning and \emph{Argument Generation} are only useful when further reasoning is required to infer a prediction, making them auxiliary to more common zero-shot approaches. Finally, we demonstrate that our approach forces larger gains in smaller language models, showcasing a complex relationship between model size and prompting methods in foundation models. 
\end{abstract}

\section{Introduction}

Large Language Models, including state-of-the-art models such as Llama family of LLMs \cite{touvron2023llama}, Mistral 7B \cite{jiang2023mistral}, and Phi-3 \cite{abdin2024phi3} have shown to significantly outperform previous generation of models \cite{wang2023chatgpt-sentiment} such as BERT \cite{devlin-etal-2019-bert} in several mainly classification tasks \cite{10.1145/3641289}. However, despite their seemingly human-like auto-regressive behavior, Large Language Models do not perform well when deep reasoning or analysis is required to effectively infer a prediction \cite{lee-etal-2023-large, tao2023eveval}. 

\begin{figure}[!ht]
    \centering
    \includegraphics[scale=0.6]{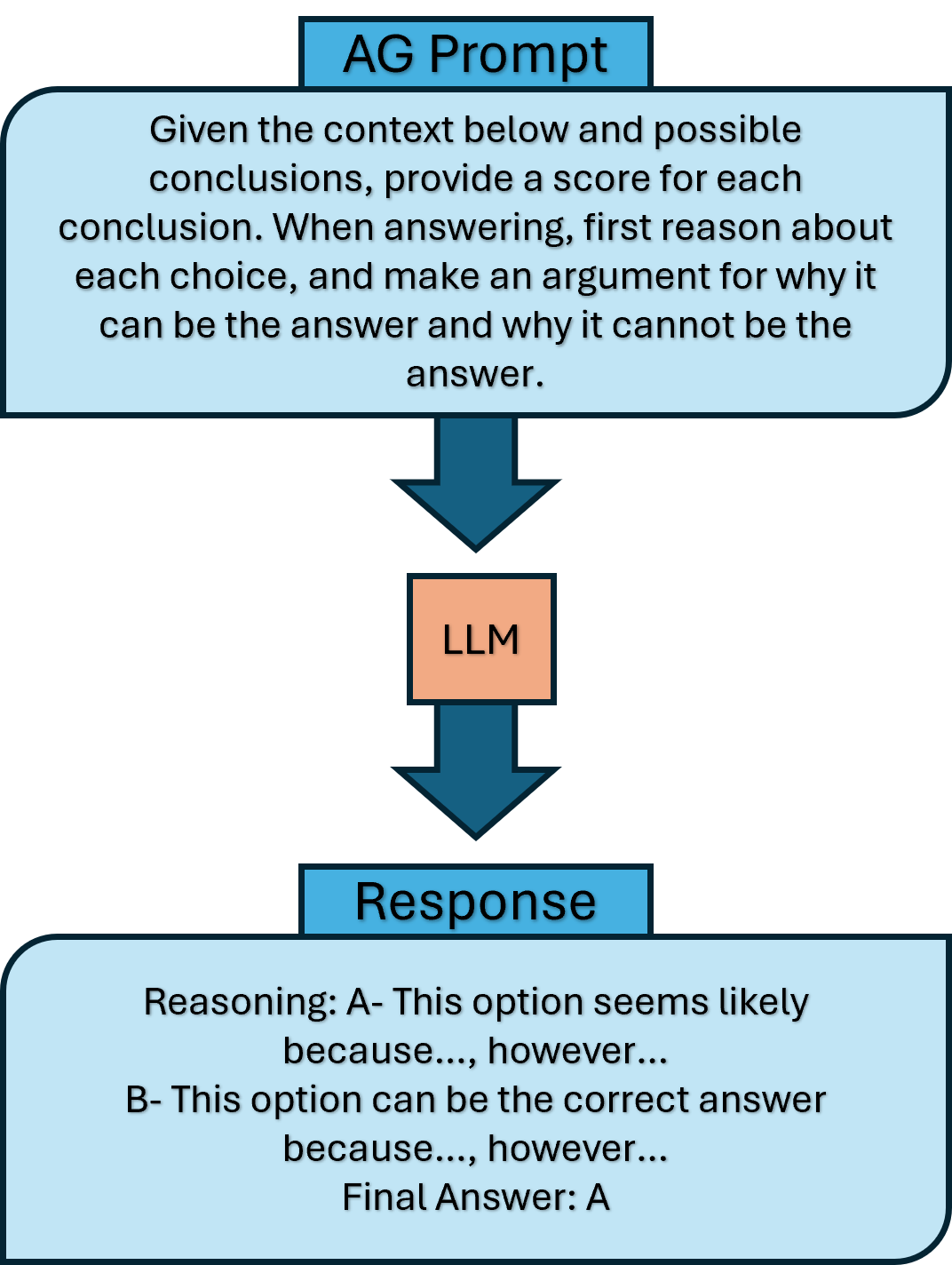}
    \caption{The general framework of Argument Generation Prompting}
    \label{framework}
\end{figure}

In order to bolster the reasoning capabilities of large language models, the research community has done extensive recent work in the form of chain-of-thought reasoning \cite{10.5555/3600270.3601883, wang2023h}, Self-Reflection \cite{NEURIPS2023_91edff07}, Multi-Agent Debate \cite{liang2023encouraging, du2023improving}, and Socratic prompting \cite{10099179}, demonstrating that prompting the model to generate the reasoning behind its answer, or generating a step-by-step guide to reach its response can help predict better results. 

Taking inspiration from chain-of-thought reasoning, and motivated by the need to develop better prompt techniques with the goal of increasing model performance in reasoning tasks, we introduce \emph{Argument Generation}, a single-pass prompting technique that aims to utilize the reasoning and argumentation capabilities of Large Language Models to generate better responses where deeper consideration of logic or reasoning is required to infer the correct result. \emph{Argument Generation} involves a two-step process. We first prompt the model to generate possible reasoning for the truthfulness of each possible option, and then we ask the model to rank the generated arguments and map its ranking to a final output in accordance with the task expectations. 

We evaluate our method on a number of openly available state-of-the-art Large Language Models using nine tasks of different natures. We find that \emph{Argument Generation} at its weakest, does not perform significantly worse than chain-of-thought reasoning, and is able to outperform both zero-shot reasoning and chain-of-thought reasoning when a deeper understanding of the task options is required. Furthermore, we note that in comparison to chain-of-thought reasoning, \emph{Argument Generation} can be used as a stronger knowledge probing technique that is useful in instances where such probing is essential, or some level of prior knowledge regarding the task is present (such as possible response candidate). However, our method does not necessarily increase the model performance for inputs that observe acceptable results under more common methods.

We make the following contributions: (1) We introduce \emph{Argument Generation}, a novel prompting technique that aims to access the underlying reasoning capabilities of LLMs. (2) We show through a series of experiments that our method is able to effectively reason under conditions that fail chain-of-thought reasoning. (3) We show that our prompting method is more effective when used with smaller language models, eliciting further investigation into the relationship between prompting approaches and model capabilities.


\section{Background and Motivation}

Argumentation is the cognitive capability of generating and evaluating ``reasons'' for deriving a conclusion \cite{Mercier_2016}.  It is a central aspect of human intelligence and is omnipresent in natural human communication. It extends the conception of reasoning in LLM-research \cite{Yu_Zhang_Tiwari_Wang_2023} by including the notion that conclusions drawn must be new. Indeed, it has been suggested that human reasoning evolved for the purposes of enabling humans to persuade each other \cite{Mercier_Sperber_2011} through arguments.

We hypothesize that many day-to-day arguments are evaluated by humans in an intuitive (fast, system 1) manner, without deep thought or ``epistemic vigilance''  \cite{Sperber_Clement_Heintz_Mascaro_Mercier_Origgi_Wilson_2010}, unless they are from trusted sources and appear to contradict our own beliefs.  Thus, because LLMs were pretrained with human communicative interactions, we hypothesize that LLMs are capable of fast argumentative thinking. By triggering argumentative thought, we hypothesize that LLMs can effectively generate reasons and assess conclusions, as well as improve core reasoning capabilities across a variety of domains, including commonsense, logical, and social.

\section{Related Work}





\paragraph{General argumentation ability of LLMs} have begun to be explored by researchers, with a focus on a number of computational argumentation subtasks such as argument mining, claim detection, evidence detection and type classification, argument generation, and summarization \cite{Balikas_2023,Chen_Cheng_Tuan_Bing_2023,Holtermann_Lauscher_Ponzetto_2022,Ruiz-Dolz_Lawrence_2023,Thorburn_Kruger,de_Wynter_Yuan_2023}. Research suggests that LLMs ``exhibit commendable performance'' \cite{Chen_Cheng_Tuan_Bing_2023} in zero-shot and few-shot settings  thereby supplying a foundation supporting our approach.

Delving deeper, we can explore two core aspects of argumentation. First, the ability to argue for/against all sides (thinking like a lawyer). Second, the ability to generate implicit assumptions (necessary or sufficient warrants) needed to support the argument.

\paragraph{Arguing all sides} is related to ``backward reasoning'' suggested  in \cite{Yu_Zhang_Tiwari_Wang_2023}, where they discuss that it is ``better to collect both supportive and opposing knowledge to compare the confidence of different conclusions for defeasible reasoning.'' Additionally \cite{Wang_Wei_Schuurmans_Le_Chi_Narang_Chowdhery_Zhou_2022} discuss the idea of allowing several different reasoning paths and choosing the ``most consistent one''. Another approach is contrastive chain-of-thought \cite{Chia_Chen_Tuan_Poria_Bing_2023} where they consider both valid and invalid reasoning demonstrations alongside original prompt -- a dual perspective approach. Additionally, work in multiagent debate, for example \cite{Chia_Chen_Tuan_Poria_Bing_2023} uses a notion of a debate with multiple agents discussing and talking about the problem. However, none of these  approaches attempt at \textit{rationalizing} all sides of an argument. That is none of these offer up the best possible argument for/against each choice, and then evaluate the best argument (for example, anticipatory reflection of plans in \cite{Wang_Li_Deng_Roth_Li_2024}).  

\paragraph{Extracting implicit information} relates to work in ``knowledge-enhanced'' \cite{qiao-etal-2023-reasoning} strategies in which an implicit model generates knowledge and rationales. Also \citet{Yu_Zhang_Tiwari_Wang_2023} discusses  Leap-of-thought reasoning which uses implicit facts to answer questions. A related notion is that of decomposing implicit multi-hop questions down in connection with the general backward reasoning tactic of question-decomposition (see summary in \cite{Yu_Zhang_Tiwari_Wang_2023}). Work by \cite{Sarathy_Burstein_Friedman_Bobrow_Kuter_2022} suggests extracting implicit assumptions from premise-conclusion pairs, however, that work does not explore how such endeavor influences an LLM's reasoning capability. Although there is a growing body of work in question decomposition, it is unclear to what extent they take implicit assumptions into account.


\paragraph{General LLM reasoning capabilities} have been improving over the past several years with numerous datasets targeting different types of reasoning -- logical, mathematical, commonsense, argumentation, and social reasoning \cite{qiao-etal-2023-reasoning,Yu_Zhang_Tiwari_Wang_2023,Huang_Chang_2023,Yu_He_Wu_Dai_Chen_2023,Luo_Kumbhar_shen_Parmar_Varshney_Banerjee_Aditya_Baral_2023,Sahoo_Singh_Saha_Jain_Mondal_Chadha_2024}. The methods have involved various techniques to evoke reasoning processes such as having the LLM explicate its chain of thought \cite{Wei_Wang_Schuurmans_Bosma_Chi_Le_Zhou_2022}, reflect on its own reasoning process \cite{Wang_Zhao_2023}, decompose complex reasoning processes into simpler problems that can be solved more easily \cite{Khot_Trivedi_Finlayson_Fu_Richardson_Clark_Sabharwal_2023}, explore many different reasoning paths and decide on one that wins a majority vote \cite{Wang_Wei_Schuurmans_Le_Chi_Narang_Chowdhery_Zhou_2022}, and others. These various methods have shown improvements in various reasoning tasks, but none have shown cross-domain effectiveness.  Moreover, their reasoning capabilities are limited when exposed to scenarios in which the model must resolve a disagreement \cite{lee-etal-2023-large}, distinguish a correct phrase from an incorrect one \cite{riccardi2023word}, or assign a nondeterministic gender to a subject \cite{zakizadeh-etal-2023-difair}. Overall, Large Language Models have shown promising results in a variety of reasoning tasks while serious challenges and shortcomings still remain \cite{10.1145/3641289}. What is missing is a cross-domain strategy to improve an LLM's zero-shot reasoning capabilities, which we hypothesize to be enhanced by its latent capability for argumentative thinking.

\section{Methodology}

We now provide details regarding our approach, including the proposed zero-shot approach and the reasoning behind our choice of \emph{Argument Generation} as a prompting technique.

\paragraph{Argument Generation} involves two overall steps. Given an initial input $x$ with possible answers $k_1, k_2,...k_n$, we first prompt the model to generate arguments supporting and attacking each answer $k_i$, creating arguments $x'_1, x'_2, ... x'_n$ for each possible answer. We then ask the model to choose the answer with the strongest argument as the final output. More concretely, the Large Language Model is utilized as a proxy for an argument ranking function that chooses the most feasible options among arguments $x'_1, x'_2, ... x'_n$. 

The rationale behind our approach is two-fold. First, it has been shown that Large Language Models, when provided with a reasoning context towards the correct output, observe significantly improved performance \cite{10.5555/3600270.3602070, 10.5555/3600270.3601883}, making the reasoning behind each choice an important contributor to model performance. Second, Large Language Models can act as effective rankers when provided with a list-wise input of possible options \cite{Ma2023ZeroShotLD}, indicating the feasibility of their possible utilization for the effective ranking of arguments. As a result, the proposed technique relies on the assumption that the correct answer $k_i$ to the query $x$ should logically have the strongest argument supporting it, forcing the ranker model to choose the argument that is directly mapped to the correct answer.

Essentially, \emph{Argument Generation} is similar to chain-of-thought reasoning because both focus on the generation of a token chain with the goal of increasing the probability of generating a viable final answer. However, chain-of-thought reasoning operates under the assumption that the generation of supporting steps is sufficient for the final true output. On the other hand, \emph{Argument Generation} aims to take into consideration the possibility of the presence of a counterargument that is statistically more significant than the answer that is generated by pure chain-of-thought. As such, we hypothesize that chain-of-thought can sufficiently generate the most logically intuitive response to the user input, while \emph{Argument Generation} might be better suited for cases where the correct answer is initially unintuitive but may increase in statistical significance as a valid counterargument is presented against the other answer candidates.

\section{Evaluation}
\label{evaluation_init}

To empirically evaluate the effectiveness of our proposed method, we have tested the performance of \emph{Argument Generation} in nine datasets and across nine models. For the remainder of this section, we focus on describing our evaluation setting. 

\subsection{Models} In order to perform a comprehensive evaluation over models of different size and architecture, we test our approach using nine models, including two families of models, and five independent, recently released LLMs. These include Llama 3 family of models (8B and 70B), Gemma family of models (2B and 7B) \cite{Mesnard2024GemmaOM}, Phi-3 3.8B \cite{abdin2024phi3}, Mistral 7B \cite{jiang2023mistral}, GPT 4o-mini\footnote{\href{https://openai.com/index/gpt-4o-mini-advancing-cost-efficient-intelligence/}{OpenAI}}, Qwen2 1.5B \cite{yang2024qwen2technicalreport}, and Aya 35B \cite{ustun-etal-2024-aya}.

\subsection{Datasets} Our choice of datasets includes candidates from nine different tasks, each representing a group of tasks that aim to quantify a specific aspect of a given model. We strive to cover tasks belonging to different domains, including question-answering, argumentation, reasoning, bias evaluation, human-alignment, and autoregressive generation. The tested datasets include CommonSenseQA \cite{talmor-etal-2019-commonsenseqa}, DiFair \cite{zakizadeh-etal-2023-difair}, IBM-30K \cite{Gretz_Friedman_Cohen-Karlik_Toledo_Lahav_Aharonov_Slonim_2020} TruthfulQA (generation and multiple choice tasks) \cite{lin-etal-2022-truthfulqa}, StereoSet \cite{nadeem-etal-2021-stereoset}, StrategyQA \cite{geva-etal-2021-aristotle}, Formal Fallacies \cite{suzgun-etal-2023-challenging}, and AlpacaEval (human annotation task) \cite{dubois2024length}. 

For all tasks, we report the metric proposed by the task's respective paper. The only exceptions to this rule are IBM-30K and the generation task of TruthfulQA. For IBM-30K, we report $1-MAE$ as the final score to be consistent with others metrics and to showcase the model response quality per individual instance. In the case of TruthfulQA, we use GPT 4o-mini as the judge model as opposed to the fine-tuned GPT-3 utilized by the authors. For the multi-choice TruthfulQA task, we additionally generate 60 questions by randomly sampling 15\% of the original dataset and replacing the correct option with `None of the Answers are Correct'. This is done in order to further evaluate model performance when no clear answer exists.

Observe that \emph{Argument Generation} requires the existence of valid candidate responses in order to correctly reason, and choose a response. However, in the case of Large Language Models, it is often the case that the user does not have a set of candidate responses for their question. In such cases, we prompt the model to generate such responses first, and then use them as the possible answers to the question. This approach is based on the hypothesis that if a model has sufficient knowledge to answer a question, it should also generate that response as a candidate. Similar methods have shown to be effective in prompt ranking approaches \cite{hu-etal-2024-rankprompt}.

\subsection{Argument Generation}

We perform our evaluations using two different \emph{Argument Generation} settings in order to evaluate both the effect of generation of \textbf{implicit assumptions}, as well as the model sensitivity to different \emph{Argument Generation} prompts. In the first approach, given an input $x$ and a possible answer $k$, we explicitly ask the model to generate an \textbf{implicit assumption} under which $k$ is a valid response to $x$. An implicit assumption is a set of logical propositions $P$ such that every proposition in $P$ must hold in order for the answer to follow logically from $x$. We then ask the model to rank these implicit assumptions by the feasibility of all $p_i \in P$ to hold simultaneously. We finally take the implicit assumption with the highest feasibility ranking as the final answer to the input. 

\begin{algorithm}
\caption{Argument Generation}
\label{main_algorithm}
\begin{algorithmic}[1]

\Require Input $x$, List of Possible Answers $K$
\Ensure Final Response $k_i$

\Procedure{Generation}{x, K}

\Function{ImplicitAssumption}{x, K}
    \State Let $A := $ \o
    \ForAll {$k_i \in K$}
        \State $A := A$ $\cup$ \Call{Assumption}{x, $k_i$}
    \EndFor
    \State Let $Ranking := $ \Call{Ranking}{A}
    \State \Return $Ranking[0]$ \Comment{Return the Top Ranking Answer}
\EndFunction

\Function{ArgumentGeneration}{x, K}

    \State Let $A := $ \o
    \ForAll {$k_i \in K$}
        \State $A := A$ $\cup$ \{\Call{Argument}{x, $k_i$}, \Call{Argument}{x, $\neg k_i$}\}
    \EndFor
    \State Let $Ranking :=$ \Call{LWR}{A}
    \State \Return $Ranking[0]$ \Comment{Return the Top Ranking Answer}
\EndFunction
\EndProcedure
\end{algorithmic}
\end{algorithm}

In the second approach, given an input $x$ and a possible answer $k$, we ask the model to both generate an argument for accepting $k$ as a correct answer to $x$ and generate an argument for rejecting $k$ as a correct answer to $x$. We then apply this process to all candidate answers $k_1$ through $k_n$ such that $n$ tuples of arguments are generated by the model. We finally prompt the model to rank the aforementioned $n$ tuples and generate the final answer to input $x$. 

Algorithm \ref{main_algorithm} showcases both of the aforementioned techniques, where \Call{Assumption}{x, K} refers to the generation of implicit assumptions for each candidate answer, and ranking them via a list-wise ranking technique, and \Call{Argument}{x, K} refers to the generation of tuples of arguments for each candidate answer that both support and attack the corresponding candidate answer, and then ranking them via a list-wise ranking approach.

We acknowledge that it is possible to extend our approach to a multi-agent setting, where the argument generation is done by an external model that is separate from the ranking model. However, we focus on single-pass prompting for the purpose of this study to (i) provide a single-pass, easy-to-implement approach that is comparable to zero-shot chain-of-thought reasoning both in performance, and running time, and (ii) refrain from unnecessarily increasing the computational requirement of the approach, as seen in other multi-agent techniques. However, we hypothesize that generalizing our algorithm to utilize multiple agents is both simple and observes an increase in performance.

\section{Evaluation Results}


\begin{table*}
\resizebox{1\textwidth}{!}{%
\setlength{\tabcolsep}{12pt}
\begin{tabular}{@{}lllllllllll@{}}
\toprule
Model & Prompt & CommonSenseQA & DiFair & IBM-30K & TruthfulQA & StereoSet & StrategyQA & TruthfulQA Gen & FormalFallacies & AlpacaEval \\ \midrule
\multirow{4}{*}{Gemma 2B} & Zero-Shot & \textbf{43.24}\% & 0.0\% & 59.46\% & 20.63\% & \textbf{63.70}\% & \textbf{55.45}\% & 34.66\% & 53.20\% & 54.39\% \\
\multicolumn{1}{c}{} & Chain of Thought & 41.85\% & 12.65\% & 49.98\% & 18.61\% & 36.17\% & 49.34\% & \textbf{34.77}\% & 53.20\% & 57.01\% \\
\multicolumn{1}{c}{} & Argument Generation w/ Implicit Assumptions & 37.18\% & 34.54\% & 62.63\% & \textbf{47.97}\% & 44.97\% & 46.28\% & 29.32\% & 49.60\% & \textbf{57.78}\% \\
\multicolumn{1}{c}{} & Argument Generation & 39.80\% & \textbf{55.39}\% & \textbf{80.93}\% & 31.27\% & 34.3\% & 50.21\% & 29.32\% & \textbf{53.60}\% & 57.62\% \\ \midrule
\multirow{4}{*}{Gemma 7B} & Zero-Shot & \textbf{69.28}\% & 0.0\% & 70.85\% & 28.93\% & \textbf{88.87}\% & \textbf{66.37}\% & 55.99\% & \textbf{49.60}\% & \textbf{62.71}\% \\
 & Chain of Thought & 69.12\% & 32.52\% & 63.14\% & \textbf{41.48}\% & 66.98\% & 58.07\% & 50.69\% & 47.20\% & 61.94\% \\
 & Argument Generation w/ Implicit Assumptions & 66.33\% & 47.51\% & 69.31\% & 33.05\% & 64.05\% & 61.33\% & 59.59\% & 47.20\% & 59.93\% \\
 & Argument Generation & 66.66\% & \textbf{55.84}\% & \textbf{72.94}\% & 25.21\% & 73.88\% & 54.14\% & \textbf{59.65}\% & 49.20\% & 57.62\% \\ \midrule
\multirow{4}{*}{Llama3 8B} & Zero-Shot & 71.33\% & 22.19\% & 60.51\% & 47.97\% & 42.47\% & 65.93\% & 47.52\% & 53.20\% & \textbf{58.24}\%  \\
 & Chain of Thought & \textbf{71.41}\% & 10.80\% & 66.03\% & 44.57\% & 54.36\% & \textbf{74.23}\% & 64.65\% & \textbf{59.20}\% & 55.00\% \\
 & Argument Generation w/ Implicit Assumptions & 63.22\% & 55.88\% & 71.22\% & \textbf{51.70}\% & \textbf{55.73}\% & 60.26\% & 78.28\% & 46.80\% & 51.30\% \\
 & Argument Generation & 64.12\% & \textbf{58.57}\% & \textbf{73.50}\% & 33.93\% & 45.90\% & 62.88\% & \textbf{78.88}\% & 50.00\% & 46.68\% \\ \midrule
\multirow{4}{*}{Llama3 70B} & Zero-Shot & 79.85\% & 78.08\% & 76.04\% & 69.04\% & 41.91\% & \textbf{72.77}\% & 57.09\% & 53.20\% & \textbf{52.22}\% \\
 & Chain of Thought & \textbf{80.26}\% & \textbf{82.79}\% & 64.46\% & \textbf{70.53\%} & 39.04\% & 74.67\% & 77.80\% & \textbf{71.60}\% & 49.36\% \\
 & Argument Generation w/ Implicit Assumptions & 74.44\% & 72.45\% & \textbf{76.98}\% & 56.91\% & \textbf{73.44}\% & 45.41\% & 82.58\% & 64.40\% & 49.52\% \\
 & Argument Generation & 75.34\% & 79.16\% & 76.13\% & 68.93\% & 52.05\% & 72.05\% & \textbf{82.59}\% & 62.80\% & 50.15\% \\ \midrule 
\multirow{4}{*}{Phi3 3.8B} & Zero-Shot & \textbf{67.97}\% & 6\% & 63.04\% & 47.55\% & 56.0\% & 64.19\% & 57.33\% & 53.20\% & 62.22\% \\
 & Chain of Thought & 66.66\% & \textbf{71.59}\% & 62.57\% & 51.48\% & 61.52\% & \textbf{64.62}\% & 63.94\% & 54.80\% & 61.63\% \\
 & Argument Generation w/ Implicit Assumptions & 66.91\% & 57.24\% & \textbf{69.50}\% & 51.70\% & 61.15\% & 60.26\% & 73.08\% & 54.80\% & \textbf{63.17}\% \\
 & Argument Generation & \textbf{67.97}\% & 52.39\% & 69.17\% & \textbf{52.12}\% & \textbf{61.67}\% & 62.88\% & \textbf{73.54}\% & \textbf{55.60}\% & 57.62\% \\ \midrule
\multirow{4}{*}{Mistral 7B} & Zero-Shot & 67.81\% & 45.66\% & 64.83\% & 8\% & 46.61\% & 61.57\% & 65.74\% & \textbf{53.20}\% & 59.93\% \\
 & Chain of Thought & \textbf{67.89}\% & 62.19\% & 59.82\% & \textbf{55.95}\% & 41.10\% & \textbf{65.06}\% & \textbf{77.91}\% & 47.20\% & \textbf{61.01}\% \\
 & Argument Generation w/ Implicit Assumptions & 64.29\% & 63.44\% & 66.58\% & 50.63\% & 46.28\% & 60.26\% & 77.29\% & 50.40\% & 58.08\% \\
 & Argument Generation & 64.70\% & \textbf{66.51}\% & \textbf{66.85}\% & 51.27\% & \textbf{49.24}\% & 60.69\%  & 77.50\% & 50.00\% & 54.54\% \\ \midrule

\multirow{4}{*}{GPT-4o-Mini} & Zero-Shot & 82.47\% & \textbf{83.58}\% & 55.78\% & \textbf{66.06}\% & 75.48\% & \textbf{77.50}\% & 66.15\% & 53.20\% & \textbf{65.63}\% \\ 
& Chain of Thought & \textbf{82.71}\% & 79.92\% & 51.25\% & 65.53\% & 86.37\% & \textbf{77.50}\% & 82.30\% & 63.20\% & 63.63\% \\ 
& Argument Generation w/ Implicit Assumptions & 79.68\% & 73.15\% & \textbf{71.96}\% & 58.29\%  & 86.22\% & 70.30\% & 91.83\% & \textbf{71.20}\% & 56.70\% \\
& Argument Generation & 80.26\% & 81.10\% & 71.71\% & 56.38\% & \textbf{86.87}\% & 71.61\% & \textbf{91.89}\% & 69.20\% & 53.77\% \\ \midrule

\multirow{4}{*}{Qwen2 1.5B} & Zero-Shot & \textbf{69.45}\% & 10.21\% & 76.04\% & 29.14\% & \textbf{50.31}\% & 54.58\% & 42.37\% & \textbf{53.20}\% & 53.15\% \\
& Chain of Thought & 59.95\% & 22.56\% & 64.46\% & \textbf{32.65}\% & 39.55\% & 54.58\% & \textbf{53.76}\% & 46.40\% & 61.32\% \\ 
& Argument Generation w/ Implicit Assumptions & 49.95\% & 50.03\% & \textbf{76.98}\% & 11.48\% & 26.99\% & 49.34\% & 44.46\% & 49.60\% & \textbf{63.02}\% \\
& Argument Generation & 54.79\% & \textbf{52.57}\% & 76.13\% & 14.68\% & 31.87\$ & \textbf{55.02}\% & 43.80\% & 50.00\% & 62.40\% \\ \midrule

\multirow{4}{*}{Aya 35B} & Zero-Shot & \textbf{85.83\%} & 69.71\% & 62.73\% & 48.82\% & \textbf{65.28}\% & 67.68\% & \textbf{44.44}\% & \textbf{53.20}\% & 65.48\% \\
& Chain of Thought & 82.39\% & \textbf{74.02}\% & 40.06\% & 43.82\% & 48.61\% & \textbf{82.53}\% & 41.81\% & 47.60\% & 63.02\% \\
& Argument Generation w/ Implicit Assumptions & 76.16\% & 61.63\% & \textbf{72.64}\% & \textbf{58.19}\% & 47.33\% & 72.48\% & 30.20\% & 48.40\% & 63.63\% \\ 
& Argument Generation & 77.31\% & 66.25\% & 64.56\% & 54.25\% & 47.85\% & 78.60\% & 29.84\% & 47.60\% & \textbf{66.10}\%
\end{tabular}%
}

\caption{Prompting results using Argument Generation, Chain of Thought Reasoning, and Zero-Shot Prompting in nine different tasks.}
\label{tab:results}
\end{table*}

We now showcase our results as tested against the datasets mentioned in section \ref{evaluation_init}. We additionally show that \emph{Argument Generation}, when outperforming zero-shot chain-of-thought reasoning, demonstrates significantly higher performance gain, and suffers smaller losses in cases where it does not result in increased performance. We finally provide a model size analysis to better understand the relationship between prompting methods and the number of parameters present in a given Large Language Model. 

\subsection{Performance Analysis}

Table \ref{tab:results} showcases the evaluation results when using \emph{Argument Generation} against zero-shot chain-of-thought prompting \cite{10.5555/3600270.3601883} and common zero-shot prompting \cite{radford2019language}. 

We observe that our method is able to outperform both zero-shot prompting and chain-of-thought reasoning in 38 of the 81 test settings, amounting to a win rate of 46.91\%. Additionally, our approach outperforms chain-of-thought reasoning in 47 of the 81 settings, showcasing that \emph{Argument Generation} yields better results in 58.02\% of the test cases. Among the 45 cases where our proposed method performs better, there are 35 cases (77.77\%) in which both proposed approaches outperform chain-of-thought reasoning, while \emph{Argument Generation} with implicit assumptions is able to yield better results in 38 cases (84.44\%), and \emph{Argument Generation} without implicit assumptions has a better performance in 42 cases (93.33\%), showcasing that both methods have similar results while tested against chain-of-thought reasoning.

With respect to individual datasets, we find that our method enjoys a significant performance boost when tested against instances of  IBM-30K \cite{Gretz_Friedman_Cohen-Karlik_Toledo_Lahav_Aharonov_Slonim_2020}, with both methods showing improved results over the two other baselines in all models. This behavior is expected as IBM-30K measures a model's capability to correctly discern a valid argument from an invalid one, and our approach operates via generating arguments that both support and attack the given input, meaning that invalid arguments will have weaker support, allowing the model to effectively rank the inputs based on their argumentative strength. 

Additionally, we observe that \emph{Argument Generation} is able to increase model performance for 10 out of 18 instances (55.55\%) against all methods, and for 13 out of 18 instances (72.22\%) against chain-of-thought reasoning in DiFair \cite{zakizadeh-etal-2023-difair} and StereoSet \cite{nadeem-etal-2021-stereoset} datasets, showcasing that argumentation might serve as a reliable debiasing method for Large Language Models. Interestingly, the correlation between our approach's improving effects and a given model's general capability is not strictly positive in this case, meaning that it is possible for larger models to observe lower, or no gains when prompted with \emph{Argument Generation}. We attribute this observation to the possibility of more capable models deceiving themselves via supporting an incorrect candidate when the initial knowledge is sufficient to make a prediction, meaning that \emph{Argument Generation} might force an artificial and unwanted decrease in model confidence. We provide further details and analysis in section \ref{model_size}.


\subsection{Performance Difference Analysis}

In order to observe the expected performance metric difference, we define $\Delta_{min}$ as the mean difference between chain-of-thought reasoning and the worst-performing \emph{Argument Generation} method when chain-of-thought reasoning is performing better than our approach, and $\Delta_{max}$ as the mean difference between chain-of-thought reasoning and the best-performing \emph{Argument Generation} method when chain-of-thought reasoning is performing better than our approach. Conversely, we define $\Gamma_{min}$ and $\Gamma_{max}$ similarly for cases in which \emph{Argument Generation} is performing better than chain-of-thought reasoning. More concretely, $\Delta$ values show the performance decrease of \emph{Argument Generation} with respect to chain-of-thought reasoning when the second approach is able to outperform our method, while $\Gamma$ values demonstrate the performance increase when \emph{Argument Generation} produces better results in comparison to chain-of-thought reasoning.

\begin{table}[!h]
\begin{tabular}{@{}lllll@{}}
\toprule
Model Name & $\Delta_{min}$ & $\Delta_{max}$ & $\Gamma_{min}$ & $\Gamma_{max}$ \\ \midrule
Gemma 2B   & 5.06           & 3.75           & \textbf{11.95} & \textbf{25.95} \\
Gemma 7B   & 7.79           & 4.30           & \textbf{10.02} & \textbf{14.02} \\
Llama3 8B  & 10.72           & 7.88           & \textbf{21.29} & \textbf{23.15} \\
Llama3 70B & \textbf{13.56}           & 3.99           & \textbf{7.40} & 13.12 \\
Phi3 3.8B  & \textbf{11.78}  & \textbf{8.04} & 4.05           & 4.62           \\
Mistral 7B & 4.16           & 3.11           & \textbf{3.99}  & \textbf{5.67}  \\
GPT-4o-Mini & 8.39 & 5.45 & \textbf{17.08} & \textbf{17.82} \\
Qwen2 1.5B & \textbf{13.42} & 10.02 & \textbf{10.85} & 11.95 \\ 
Aya 35B & 8.38 & 5.83 & \textbf{11.84} & \textbf{16.67} \\
\midrule
Overall    & 10.33           & 7.03          & \textbf{11.48} & \textbf{15.35} \\ \bottomrule
\end{tabular}
\caption{Observed results of $\Delta_{min}$, $\Delta_{max}$, $\Gamma_{min}$, and $\Gamma_{max}$ for all tested models. We find that in cases where our method performs better, it generally holds that it has a larger performance gain in comparison to the instances where Chain-of-Thought reasoning is the best method.}
\label{tab:delta_analysis}
\end{table}

Table \ref{tab:delta_analysis} showcases our empirical results. We find that except for the Phi3 3.8B model, all LLMs demonstrate significantly higher performance in instances where our method outperforms zero-shot chain-of-thought reasoning. Most significantly, Llama3 8B has a mean performance difference of 21.29\% between the worst-performing \emph{Argument Generation} approach and zero-shot chain-of-thought reasoning ($\Gamma_{min})$ in tasks that our method performs better. Looking at $\Gamma_{max}$, the best-performing proposed method is able to boost Gemma 2B model performance by 25.95\%, and Llama3 8B performance by 23.15\%, showcasing that overall when such an increase in model performance is observed, the increase is significant. Conversely, Phi3 3.8B, when prompted using our method, only has an increased output value of 4.62\% at best, while performing 11.78\% better than the worst-performing \emph{Argument Generation} approach, and 8.04\% better than the best-performing approach in instances that chain-of-thought reasoning yields better results. We attribute this behavior to the model's lower argument ranking capabilities, meaning that Phi3 cannot effectively rank the arguments based on their validity. This notion is further bolstered by the model's relatively low performance in the IBM-30K task when using our proposed method, as seen in Table \ref{tab:results}. Additionally, Phi family of models enjoy a significant performance boost when paired with chain-of-thought reasoning \footnote{\href{https://huggingface.co/spaces/logikon/open_cot_leaderboard}{Open COT Leaderboard}}, which we believe contributes to the observation that our approach does not significantly increase the model performance in this instance relative to other models. Overall, our observations suggest that the effectiveness of prompting techniques might be as much model-dependent as they are task-dependent.

Finally, in order to better understand the model sensitivity to the presence or absence of implicit assumptions in the designed prompts, we report the average performance difference between the two \emph{Argument Generation} methods. We find an absolute performance difference of 4.09\% between the two approaches, the lowest amount among every other possible pair, with the closest pair being chain-of-thought reasoning and normal \emph{Argument Generation} with an absolute performance difference of 8.56\%. Similarly, the two \emph{Argument Generation} methods have a Spearman correlation coefficient of 0.8351, with the closest pair having a correlation coefficient of 0.6685. Overall, our tests show that different models are generally resilient to variations in the prompt design as long as they are bound by the general procedure as provided in algorithm \ref{main_algorithm}.

\subsection{Model Size Analysis} 
\label{model_size}

\begin{figure*}[hbt!]
    \centering
    \includegraphics[width = \textwidth]{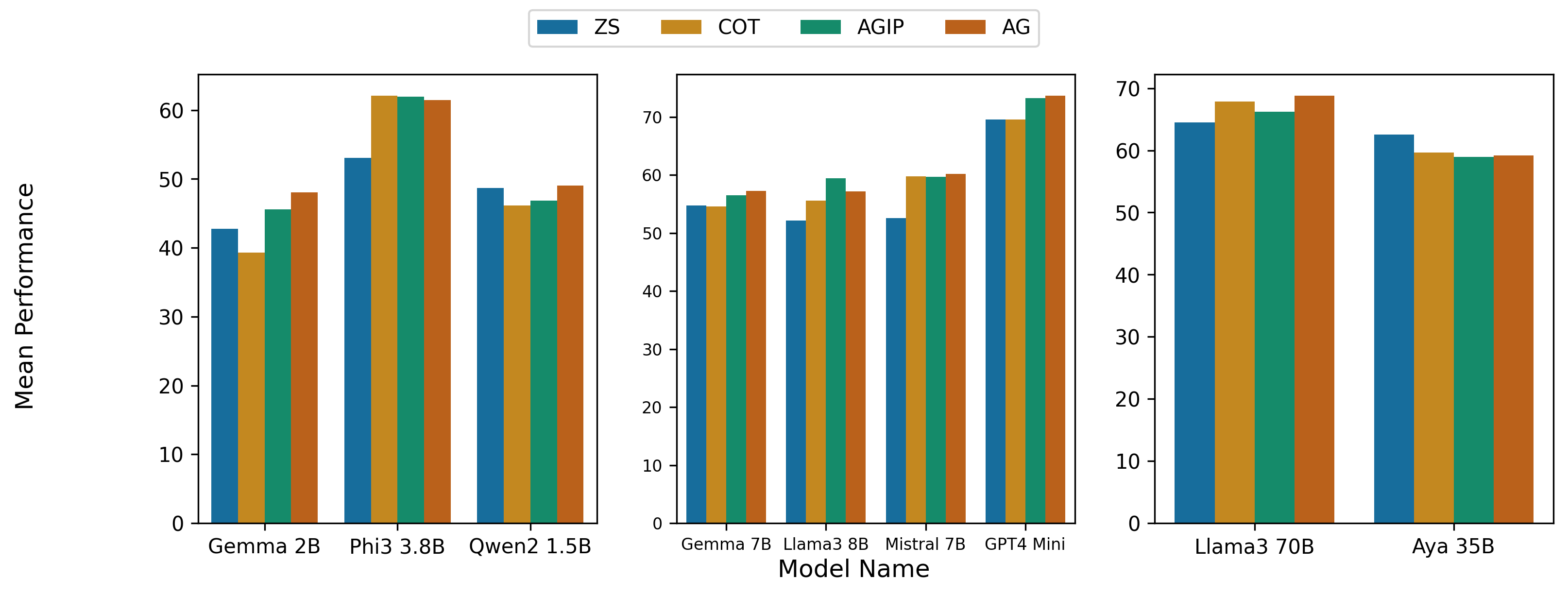}
    \caption{Mean Performance in models of different size}
    \label{fig:size_analysis}
\end{figure*}

We now provide our results on the effects of prompting on models of different sizes. In order to conduct our evaluation, we divide the models under test into three subcategories. The first category constitutes Gemma 2B, Phi3 3.8B, and Qwen2 1.5B and is demonstrative of small language models (below 7 billion parameters). The second category contains Gemma 7B, Llama3 8B, Mistral 7B, and GPT4o mini and showcases language models of medium size. Finally, Llama 3 70B and Aya 35B are members of the third category and act as sample members for the largest of language models by parameter count. 

Figure \ref{fig:size_analysis} demonstrates the mean performance of the four prompting methods across different sizes, grouped by the aforementioned categorization where ZS, COT, AGIP, and AG stand for zero-shot, chain-of-thought, Argument Generation with Implicit Assumptions, and Argument Generation, respectively. Our findings show that generally, models experience a performance increase when prompted either with chain-of-thought reasoning, or \emph{Argument Generation} with Aya 35B being the only significant exception. We observe that models of smaller sizes (medium and small) experience a significant performance boost when prompted via \emph{Argument Generation} (for 100\% of the models) and chain-of-thought reasoning (for 62\% of the models). 

Furthermore, smaller models show a higher performance gain when compared to the largest Llama 3 and Aya instances. More specifically, the mean performance gain when utilizing \emph{Argument Generation} compared to chain of thought prompting is 3.18\% for small models, and 2.72\% for medium models, while the performance gain for the large models is 0.95\%. We hypothesize that the reason behind the lower performance gain in larger models is due to their already impressive capability to infer the correct results without the requirement to introduce further information probing techniques such as chain-of-thought reasoning and \emph{Argument Generation}. More concretely, forcing the model to perform self-reasoning or rank the validity of arguments and responses does not expose the model to previously hidden information, and does not necessarily increase the performance when additional information is not strictly required to respond to the input. This phenomenon is especially observable in CommonSenseQA and TruthfulQA as seen in table \ref{tab:results}, where the introduction of prompting does not improve the model performance in all instances. These observations are in line with those reported by \citet{10.5555/3600270.3601883} and lead us to believe that knowledge probing prompting methods are only useful in cases where this additional information is required to make strong predictions and might additionally depend on model architecture. 

\begin{figure*}[hbt!]
    \centering
    \includegraphics[width = \textwidth]{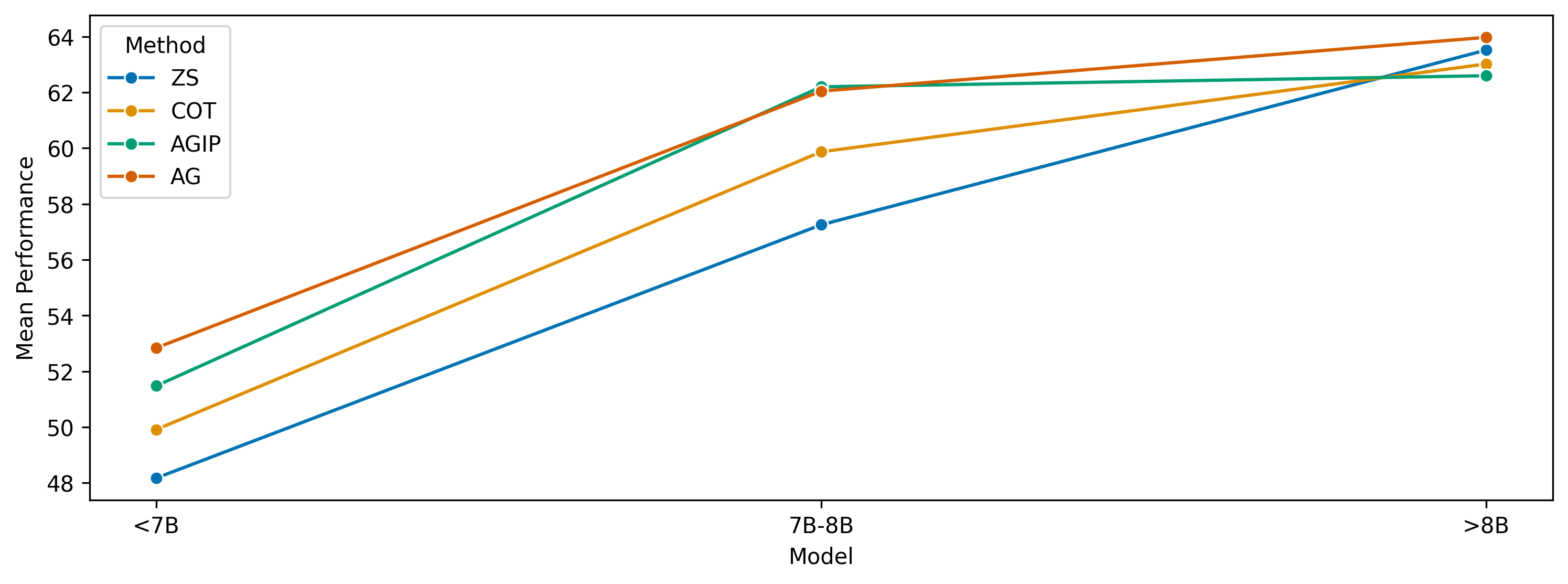}
    \caption{Mean Performance trend across model parameters}
    \label{fig:progress_analysis}
\end{figure*}

To further investigate the effects of prompting on model performance, and its relationship with the number of model parameters, we report the mean performance across the number of parameters in figure \ref{fig:progress_analysis}. We find that although both our proposed method and chain-of-thought reasoning provide improved performance in models of larger size, their impact diminishes as the models grow larger. More specifically, we find that the mean difference between zero-shot prompting and \emph{Argument Generation} methods is 4.66\% for models with less than 7 billion parameters, 4.94\% for models of 7 billion to 8 billion parameters, and 0.45\% for the largest models. Further investigation is required to fully confirm our observations, however, this finding bolsters the previous hypothesis that \emph{Argument Generation} as a prompting technique, is more effective in increasing the performance of smaller models. This behavior may stem from the fact that large models are able to generate convincing arguments for incorrect options, making the task of discerning an invalid argument from a valid one difficult. Conversely, smaller models are not able to generate arguments of high quality for incorrect candidates, thus goading the model to rank the valid argument over the incorrect one. Similarly, the observed mean differences between \emph{Argument Generation} and chain-of-thought reasoning are 2.92\%, 2.33\%, and 0.95\% respectively for models of small (<7B), medium (7B and 8B), and large (>8B) sizes.

Based on the above observation, a multi-agent technique to increase performance might be to generate arguments using a less capable model, while utilizing a more performant model to rank the arguments. We delegate these additional analyses to future work.

\section{Discussion and Future Work}

Prompting has been proposed as a method of improving model performance in either task-specific settings or broader, task-agnostic environments \cite{sahoo2024systematic}. Despite the visible gains of employing prompting to yield better model results, the literature showcasing how, and when prompting works is limited \cite{petrov2024prompting}. We observe that the proposed method is able to significantly boost the model performance in smaller models while gaining marginal improvements as the model size increases, which is contrary to the previous work showing that larger models have higher gains through prompting \cite{10.5555/3600270.3602070}. This leads us to believe that the relationship between prompting and the nature of the model is complex, and might be affected both by the model size, and its relative task-specific knowledge and capabilities. Further work is required to demonstrate the effects of prompting when models hold knowledge of varying degrees with respect to a task description. Investigation of the learning resources used in model training can provide invaluable insight into the relationship between prompting and model reasoning.

\section{Conclusion}

In this work, we have proposed \emph{Argument Generation} as a novel, zero-shot prompting technique. Through empirical evaluation using a number of datasets, we observe that our method is able to outperform both zero-shot prompting and zero-shot chain-of-thought reasoning in the majority of the conducted tests, making it a likely candidate when improving the model performance in a zero-shot setting is required. Furthermore, we show that our approach yields larger gains in smaller models, both offering an effective method that can be used in small models and providing a possible future direction to better understand the relationship between model capabilities and prompting.  

\section{Limitations}

Despite the observation that \emph{Argument Generation} is able to generally outperform other common zero-shot prompting methods, its reliance on the existence of a predefined number of options from which the model can arguments is an inherent limitation of our work. While it is true that all questions can be modified to behave as either a multi-choice question or a yes-no question, this conversion relies on the background knowledge of the user that is interacting with the model, meaning that in cases where the user has no information regarding the possible answer for an open question, the correct formulation of the input to fit our criteria can only be delegated to the model itself.

In addition, while we have made the best effort to cover datasets pertaining to different tasks that evaluate various model capabilities, it is possible that other task-agnostic prompting methods outperform our approach in a number of yet untested metrics. Further investigation is required to fully confirm the effects of our approach on different models and tasks. 

\section{Ethical Considerations}

Previous work has shown that Large Language Models are limited in their capability to understand their own lack of knowledge \cite{yin-etal-2023-large}. As such, it is possible to generate prompts that exacerbate model hallucinations, and even force models to generate misinformation. The proposed method can especially be prone to attacks of a similar kind as a malicious agent can force the model to showcase generally unwanted behavior by providing the model with incorrect, and even dangerous options. Based on this observation, we encourage the research community to continue the work in hallucination reduction and use all prompting methods both responsibly and skeptically. 

\section{Acknowledgments}

This research was supported in part by Other Transaction award HR00112490378 from the U.S. Defense Advanced Research Projects Agency (DARPA) Friction for Accountability in Conversational Transactions (FACT) program.

\bibliography{anthology, custom, argumentation, llm_reasoning}

\clearpage
\appendix
\section{Model Details}
We utilize the Ollama framework \footnote{\href{https://github.com/ollama/ollama-python}{https://github.com/ollama/ollama-python}} to conduct all evaluations described in the paper. Generally, we make use of the 4-bit quantized \cite{10.1145/3623402} versions of the tested models to maintain consistency, and due to hardware limitations. Table \ref{tab:model_source} demonstrates all the tested models, their Ollama hub links, as well as their quantization methods. In the cases that an Ollama model is not available, or the model is closed-source, we use the associated Huggingface\footnote{\href{https://huggingface.co/}{https://huggingface.co/}} instance of the model, or use an API to access the model.

\begin{table}[!htbp]
\begin{tabular}{@{}lll@{}}
\toprule
Model Name & Hub Link                                                & Quantization Method \\ \midrule
Gemma 2B   & \href{https://www.ollama.com/library/gemma:2b}{Link}    & Q4                \\
Gemma 7B   & \href{https://www.ollama.com/library/gemma:7b}{Link}    & Q4                \\
Llama3 7B  & \href{https://www.ollama.com/library/llama3:8b}{Link}   & Q4                \\
Llama3 80B & \href{https://www.ollama.com/library/llama3:70b}{Link}  & Q4                \\
Phi3 3.8B  & \href{https://ollama.com/herald/phi3-128k:latest}{Link} & Q5               \\
Mistral 7B & \href{https://www.ollama.com/library/mistral:7b}{Link}  & Q4 
    \\
GPT-4o-Mini & \href{https://huggingface.co/}{Link} & N/A 
    \\
Qwen 2 1.5B & \href{https://huggingface.co/Qwen/Qwen2-1.5B-Instruct}{Link} & FP16
    \\ 
Aya 35B & \href{https://ollama.com/library/aya:35b}{Link} & Q4
\end{tabular}
\caption{All model sources as well as their quantization method.}
\label{tab:model_source}
\end{table}

Additionally, in order to minimize output variance and generate reproducible evaluations, all tests were performed with a model temperature of 0 and a random seed of 42. Furthermore, our test setting involved a workstation containing an Nvidia A6000, and an Nvidia RTX 4090, with 128 GB of available RAM. All testing code will be made publicly available upon the publication of the work. 

\section{Evaluation Method and Prompt Strings}

Table \ref{tab:special_instructions} lists the tested prompting methods as well as the special instruction used for each prompt. A special instruction is a text string that is appended to the end of the input question and aims to guide the model behavior while responding to that specific input. 

For the case of zero-shot prompting, we simply ask the model to only respond with the correct answer without providing any instructions to reason about the input. Chain-of-thought reasoning is additionally employed via the guidelines provided by \citet{10.5555/3600270.3601883}. Finally, we showcase the special instructions for the proposed method, both containing the implicit assumption generation, and common argument generation. 

\begin{table*}[!htbp]
\begin{tabular}{p{0.35\linewidth} | p{0.6\linewidth}}
\toprule
\multicolumn{1}{c}{Prompting Method}        & \multicolumn{1}{c}{Special Instruction}   \\ \midrule
Zero-Shot                                   & Only respond with the correct answer      \\ \midrule
Chain-of-Thought                            & Let's think about each option step by step \\ \midrule
Argument Generation w/ Implicit Assumptions & When answering, first reason about each choice, and make an argument for why it can be the answer and why it cannot be the answer. Then identify, for each choice, what implicit assumptions you might be making for each of your arguments. By implicit assumption, we mean those propositions that are necessary so that the choice logically follows the question. Then select one of the choices based on the strongest argument \\ \midrule
Argument Generation & When answering, first reason about each choice, and make an argument for why it can be the answer and why it cannot be the answer. Then select one of the choices based on the strongest argument. \\ \bottomrule
\end{tabular}
\caption{Special model instructions corresponding to each prompting method.}
\label{tab:special_instructions}
\end{table*}

\end{document}